%% file: bach.tex
\documentclass[runningheads,a4paper]{llncs}

\usepackage{amssymb}
\usepackage{graphicx}
\usepackage[font=small]{caption}
\usepackage{subfiles}
\usepackage{url}
\usepackage{tabularx}
\usepackage{array,multirow}
\usepackage{siunitx}
\usepackage{hyperref}

\urldef{\mailsa}\path|{kamyar.nazeri, azad.aminpour, mehran.ebrahimi}@uoit.ca|    
\newcommand{\keywords}[1]{\par\addvspace\baselineskip
\noindent\keywordname\enspace\ignorespaces#1}

\begin{document}

\mainmatter  

\title{Two-Stage Convolutional Neural Network \\ for Breast Cancer Histology Image Classification}
\titlerunning{ }

\author{Kamyar Nazeri\and Azad Aminpour\and Mehran Ebrahimi \thanks{Corresponding author}}
\authorrunning{ }
\institute{Faculty of Science, University of Ontario Institute of Technology\\
	2000 Simcoe Street North, Oshawa, Ontario, Canada ~  L1H 7K4\\
\mailsa\\
\url{http://www.ImagingLab.ca/}
}

\maketitle

\begin{abstract}
	This paper explores the problem of breast tissue classification of microscopy images. Based on the predominant cancer type the goal is to classify images into four categories of normal, benign, in situ carcinoma, and invasive carcinoma. Given a suitable training dataset, we utilize deep learning techniques to address the classification problem. Due to the large size of each image in the training dataset, we propose a patch-based technique which consists of two consecutive convolutional neural networks. The first ``patch-wise'' network acts as an auto-encoder that extracts the most salient features of image patches while the second ``image-wise'' network performs classification of the whole image. The first network is pre-trained and aimed at extracting local information while the second network obtains global information of an input image. We trained the networks using the ICIAR 2018 grand challenge on BreAst Cancer Histology (BACH) dataset. The proposed method yields $95\%$ accuracy on the validation set compared to previously reported $77\%$ accuracy rates in the literature. Our code is publicly available at \url{https://github.com/ImagingLab/ICIAR2018}.
\end{abstract}

\section{Introduction}
\subfile{sections/introduction}

\section{Related Works}
\subfile{sections/related}

\section{Methods}
\subfile{sections/method}

\section{Experiments and Results}

\subfile{sections/results}

\section{Conclusions and Future Work}
\subfile{sections/conclusion}

\subsubsection*{Acknowledgments}
This research was supported in part by a Natural Sciences and Engineering Research Council of Canada (NSERC) Discovery Grant (DG) for ME. AA would like to acknowledge UOIT for a doctoral graduate international tuition scholarship (GITS). The authors gratefully acknowledge the support of NVIDIA Corporation for their donation of Titan XP GPU used in this research through its Academic Grant Program.

{
	\small
	\bibliographystyle{splncs}
	\bibliography{bach}
}
\end{document}

%% file: sections/introduction.tex
Breast cancer is one of the leading causes of cancer-related death in women around the world \cite{siegel2016cancer}. According to Canadian Cancer Society, over  $26,000$ women were diagnosed with breast cancer in Canada in 2017 which represents $25\%$ of all new cancer cases in women. In the same year, more than $5,000$ women in Canada lost their lives due to breast cancer which represents $13\%$ of all cancer deaths in women.

It is evident that early diagnosis can significantly increase treatment success. Breast cancer symptoms and signs are varied and diagnosis includes physical exam, mammography, ultrasound testing, and biopsy. Biopsy is generally performed after detection of some abnormality using mammography and ultrasound. 
	
	In biopsy, a sample of tissue is surgically removed to be analyzed. This can indicate which cells are cancerous, and if so which type of cancer these are associated to. Microscopy imaging data of biopsy samples are large in size and complex in nature. Therefore, pathologists face a substantial workload increase for histopathological cancer diagnosis. In recent years, the development of computer aid diagnosis (CAD) systems have helped reducing this workload. Digital pathology continues to gain momentum worldwide for diagnostic purposes \cite{ghaznavi2013digital}. 
	
	
Recently, deep learning techniques have emerged to address many problems in the field of medical image processing. We propose a classification scheme for breast cancer tissue image classification based on deep convolutional neural networks (CNN). Convolutional networks are considered state of the art technique for classification problems when the input is high-dimensional data such as images. These networks ``learn'' to extract local features from images and classify the input according to the extracted features.  Size of microscopy images are very large and due to hardware barriers, several patch-based CNN methods have been proposed in the literature \cite{araujo2017classification,hou2016patch,cruz2014automatic} to process the input image as a set of smaller patches. In these models, each image is divided into smaller patches and each patch is classified with a ``patch-wise'' classifier network and assigned to a label. To classify at the whole image level, the patch-wise network is followed by another classifier that receives output labels from the first network as input and generates label scores. These techniques achieve high accuracy with high confidence on image patches, however, they fail to capture global attributes of the image: Once all image patches are labelled, the spatial information is ignored and any possible feature that is shared between patches is lost.
	
We propose a novel two-stage convolutional neural network pipeline in a patch-wise fashion that is designed to utilize both local and global information of the input. The proposed method does not require a large memory footprint of the end-to-end training. In this scheme, the sole purpose of the patch-wise network is to extract spatially smaller feature maps from each patch. Once trained, this network is then used to extract the most salient feature maps from all patches in an image based on their local information. These feature maps are stacked together to form a spatially smaller 3D input for the ``image-wise'' network. This network is trained to classify images based on local features extracted from image and global information shared between different patches. We trained our network using the ICIAR 2018 grand challenge on BreAst Cancer Histology (BACH) dataset \cite{iciar2018} containing $400$ Hematoxylin and Eosin (H\&E) stained breast histology microscopy images. Our model has achieved 95\% accuracy on the validation set, outperforming \cite{araujo2017classification} in terms of classification accuracy. \footnote{Our code and pre-trained weights are available at \url{https://github.com/ImagingLab/ICIAR2018}.}

%% file: sections/related.tex
	Due to the importance of detection and classification of breast cancer in microscopic tissue images, many new methods have emerged in recent years \cite{araujo2017classification,hou2016patch,cruz2014automatic}. Computer aided diagnosis (CAD) systems appear to become fast and inexpensive alternatives to second opinion methods. Recently, deep learning techniques have made a huge impact in various problems including medical image processing. 
	
In the past few years, several works aimed at breast cancer detection and classification using CNNs have been published \cite{kowal2013computer,george2014remote,filipczuk2013computer,brook2006breast,zhang2011breast,cruz2014automatic,cirecsan2013mitosis}. Although the aim of all of these works are very similar, each work considers a specific type of problem. For example, \cite{kowal2013computer,george2014remote,filipczuk2013computer} are proposing a two class (malignant and benign) classifier. Other works in \cite{brook2006breast,zhang2011breast} consider more complex 3-class classification (normal, in situ carcinoma, and invasive carcinoma). Finally,  \cite{cruz2014automatic,cirecsan2013mitosis} develop a segmentation scheme for breast cancer.	
	
Our work is similar to the work of Ara{\'u}jo et al. \cite{araujo2017classification} in nature. To the best of our knowledge they were the first team to consider a four class classifier for breast tissue images. They developed a CNN followed by a  support vector machines (SVM) classifier. In their technique, first the original image is divided into twelve contiguous non-overlapping patches. The patch class probability is computed using the patch-wise trained CNN and CNN+SVM classifiers. Finally, the image-wise classification is obtained using three different patch probability fusion methods. These three methods namely include ``majority voting'', ``maximum probability'', and ``sum of probabilities'' \cite{araujo2017classification}.

%% file: sections/method.tex
Given a high resolution ($2048 \times 1536$) histology image, our goal is to classify the image into four classes: \textit{normal tissue}, \textit{benign tissue}, \textit{in situ carcinoma} and \textit{invasive carcinoma}. 
\subsection{Patch-Based Method with CNN}
The high resolution nature of the images in our dataset and the need to extract relevant discriminatory features from them impose extra limitations in implementing a regular feed forward convolutional network. Training a CNN on high resolution image requires either a very large memory footprint, which is not available in most cases, or to progressively reduce the spatial size of the image such that the downsampled version could be stored in the memory. However, downsampling an image increases the risk of losing discriminative features such as nuclei information and their densities to correctly classify carcinoma versus non-carcinoma cells. Also, if trained on the large microscopy image, the network might learn to rely  only on the most distinctive features and totally discard everything else.

\begin{figure}[hb]
	\centering
	\includegraphics[width=1\linewidth]{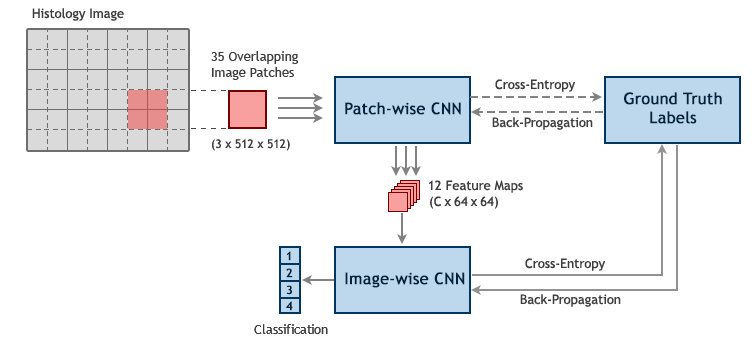}
	\caption{\small An overview of the proposed workflow. A CNN is trained on image patches. Feature maps extracted from all image patches are stacked together and passed to a second CNN that learns to predict image-level labels.}
	\label{fig:network}
\end{figure}

We follow the patch-wise CNN method proposed by \cite{araujo2017classification,hou2016patch,cruz2014automatic} followed by an image-wise CNN that classifies histology images into four classes. Given a microscopy image, we extract fixed size patches by sliding a patch (window) of size $k\times k$ with stride of $s$ over an image. This makes a total number of $[1+\frac{I_W-k}{s}] \times [1+\frac{I_H-k}{s}]$ patches where $I_W$ and $I_H$ are image width and height respectively. In our experiments, we follow \cite{araujo2017classification} and choose patch size of $k=512$ considering the amount of GPU memory available.
We also choose a stride of $s=256$, which results in $7 \times 5=35$ overlapping image patches. We argue that allowing the overlap is essential for the patch-wise network to learn features shared between patches. The proper stride $s$ is chosen by considering the receptive field of both networks when they ``work'' together, as explained later. An overview of our two-stage CNN is presented in Figure \ref{fig:network}. Note that the labels in the training set are provided only for the whole image and individual patch labels are unknown, yet we train the patch-wise network using categorical cross-entropy loss based on the label of the corresponding microscopy image. This network works like an auto-encoder that learns to extract the most salient features of image patches. Once trained, we discard the classifier layer of this network and use the last convolutional layer to extract feature maps of size $(C \times 64 \times 64)$ from any number of patches in an image, where $C$ is another hyper-parameter in our proposed system that controls the depth of output feature maps as explained later. 

To train the image-wise network, we no longer extract overlapping patches from the image: with stride of $s=512$ patches do not overlap and the total patches extracted from an image becomes $12$. We found non-overlapping patches work slightly better in our validation set. We argue that it is because overlapping patches introduce redundant features for a single concept and as a result accuracy of the image-wise network will suffer. The extracted feature-maps from all 12 patches are concatenated together to form a spatially smaller 3D input of size $(12 \times C, 64, 64)$ for the image-wise network. This network is trained against image-level labels using categorical cross-entropy loss and learns to classify images based on local features extracted from image patches and global information shared between different patches. 
\vspace{-10px}
\vspace{-20px}
\\
Once both networks are trained, we use them jointly to infer image-level class predictions.
\subsection{Network Architecture}
\begin{figure}[!htb]
	\centering
	\includegraphics[width=1\linewidth]{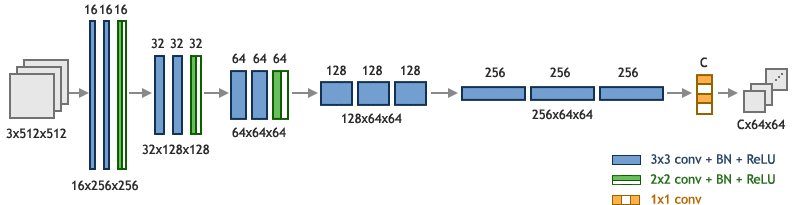}
	\caption{Patch-wise network architecture}
	\label{fig:pw_network}
\end{figure}

Inspired by \cite{simonyan2014very}, we design our patch-wise CNN using a series of $3 \times 3$ convolutional layers followed by a pooling layer with the number of channels being doubled after each downsampling. All convolutional layers are followed by \textsl{batch normalization} \cite{ioffe2015batch} and \textsl{ReLU non-linearity} \cite{maas2013rectifier}. We followed the guideline in \cite{springenberg2014striving} to implement a homogeneous fully convolutional network with occasional dimensionality reduction by using a stride of $2$. In our tests, we found that $2 \times 2$ kernel with stride of $2$ worked better than conventional max-pooling layers in terms of performance. Instead of fully connected layers for the classification task, we use a $1 \times 1$ convolutional layer to obtain the spatial average of feature maps from the convolutional layer below it, as the confidence categories and the resulting vector is fed into the \textsl{softmax layer} \cite{lin2013network}. We use this feature map later as an input to the image-wise CNN. To further control and experiment the effect of using spatial averaging layer, we introduce another hyper-parameter $C$ that controls the depth of the output feature maps. Both batch normalization and global average pooling are \textsl{structural regularizers} \cite{lin2013network,ioffe2015batch} which natively prevent overfitting. As a result, we did not introduce any dropout or weight decay in our model. Overall, there are $16$ convolutional layers in the network with the input being downsampled $3$ times at layers $3$, $6$, and $9$. Figure \ref{fig:pw_network} illustrates the overall structure of the proposed patch-wise network. 

\begin{figure}[!htb]
	\centering
	\includegraphics[width=0.9\linewidth]{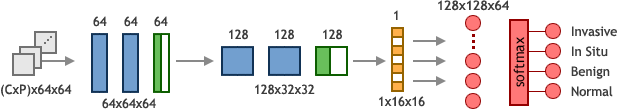}
	\caption{Image-wise network architecture}
	\label{fig:iw_network}
\end{figure}

\noindent
For the proposed image-wise network we follow a similar pattern. Series of $3 \times 3$ convolutional layers are followed by a $2 \times 2$ convolution with stride of $2$ for downsampling. Each layer is followed by batch normalization and ReLU activation function. We use the same $1 \times 1$ convolutional layer as before to obtain the spatial average of activation maps before the classifier. The convolutional layers are followed by $3$ fully connected layers with a softmax classifier at the end. Unlike the patch-wise network, overfitting is a major problem for this network, as a result, we heavily regularize this network using dropout \cite{srivastava2014dropout} with the rate of $0.5$ and use early stopping once the validation accuracy doesn't improve to limit overfitting. Network architecture is shown in Figure \ref{fig:iw_network}. 
\\ 
\noindent
The receptive field of the last convolutional layer with respect to the patch-wise network is $252$ \cite{luo2016understanding}. In principle, this number has to be the maximum stride value $s$ we can choose in extracting patches, in order to cover the whole surface of the input image. In our experiments we found $s=256$ has almost the same accuracy. We argue that having large patch size $(512\times 512)$ makes our network invariant to small changes in $s$. Note that small values of $s$ makes training time slower and the network prone to overfitting.

%% file: sections/results.tex
\begin{figure}[!htb]
		\centering
		\includegraphics[width=1\linewidth]{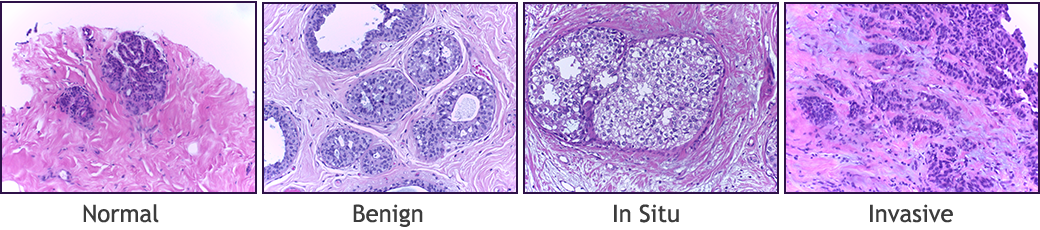}
		\caption{Samples of ICIAR 2018 challenge on BreAst Cancer Histology (BACH) dataset}
		\label{fig:dataset}
	\end{figure}

	Our dataset is composed of $400$ high resolution Hematoxylin and Eosin (H\&E) stained breast histology microscopy images labelled as \textit{normal}, \textit{benign}, \textit{in situ carcinoma}, and \textit{invasive carcinoma} ($100$ images for each category). These images are patches extracted from whole-slide images and annotated by two medical experts. Images for which there was a disagreement between pathologists were discarded. Figure \ref{fig:dataset} highlights the variability in sample images. The dataset was available at \url{https://iciar2018-challenge.grand-challenge.org/dataset/}. 
	\vspace{-15px}
	\vspace{-10px}
	\noindent
	\\The size of dataset is relatively small for our convolutional network. To prevent patch-wise network from overfitting, we apply several data augmentations on patches extracted from the microscopy image. Pathology images do not have canonical orientation and the classification problem is rotation invariant \cite{cirecsan2013mitosis,araujo2017classification}. To augment the dataset we rotate each patch by $4$ multiples of $\ang{90}$, with and without mirroring, which results in $8$ valid variations for each patch. We further apply random color perturbations to these variations as suggested by \cite{liu2017detecting} and produce $8$ more patches. The color augmentation process would help our model to learn color-invariant features and make pre-processing color normalization \cite{macenko2009method} step unnecessary. The total size of our patch-wise dataset becomes $16 \times 35 \times 400$.

	We train both networks on a single NVIDIA Titan XP GPU using Adam \cite{kingma2014adam} optimizer, mini-batch size of $64$ and initial learning rate of $0.001$, with a decay of $0.1$ every $20$ epochs. We train the patch-wise model on $80\%$ of our dataset for $30$ epochs and used the remaining $20\%$ for cross-validation. We use the same train/validation sets for image-wise network. 
	
	The accuracy of the proposed method is measured as the ratio between correct samples and the total number of evaluated images. For our validation set of $80$ images, our best model achieved $93.75\%$ accuracy (Table \ref{tab:comp}) and mean Area Under Curve (AUC) of $98.3$ corresponding to ($98.9$, $97.7$, $98.5$, $98.1$) for the four classes (Table \ref{tab:res}) based on Receiver Operating Characteristic (ROC) analysis. In addition, we experimented with an ensemble model, averaging across $8$ variations of rotation/flip in the input image as suggested by \cite{liu2017detecting} and further improved the accuracy to $95.00\%$. 
	
\begin{figure}
		\centering
		\includegraphics[width=1\linewidth]{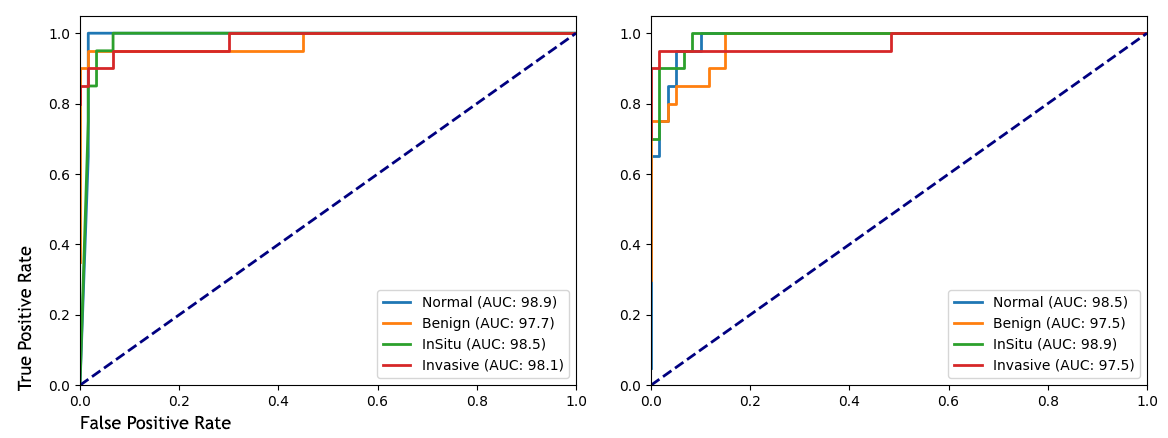}
		\caption{Receiver Operating Characteristic (ROC) curves for two output sizes of the patch-wise network. Left: 1-Channel feature map, Right: 4-Channel feature map }
		\label{fig:roc}
	\end{figure}

	To compare our results with those of Ara{\'u}jo et al.\cite{araujo2017classification}, the image-wise labels through a decision making scheme on outputs of the patch-wise network is also presented in (Table \ref{tab:comp}). The image label is obtained using one of the three different patch probability fusion methods. These methods include, \textsl{majority voting}, where the image label is selected as the most common patch label, \textsl{maximum probability}, where the patch with higher class probability decides the image label, and \textsl{sum of probabilities}, where the patch class probabilities are summed up and the class with the largest value is assigned. As shown in (Table \ref{tab:comp}), our proposed patch-wise network is outperforming previous methods by a large margin. 
	\vspace{-15 px}
	\begin{table}[!htb]
		\begin{center}
		\caption{Accuracy of the proposed method for one-channel output compared to \cite{araujo2017classification}} 
		\label{tab:comp}
		\def\arraystretch{1.3}
		\begin{tabular}{c|c c c|c c|}
			\cline{2-6}
		    &\multicolumn{3}{c|}{Patch-wise} & \multicolumn{2}{c|}{Image-wise}\\
		  	\cline{2-6}
		 	& Sum & Max. & Maj. & ~CNN & Ensemble \\ \hline
		  	\multicolumn{1}{|c|}{Proposed} & 91.25 & 92.50 & 90.00 & ~93.75 & \textbf{95.00} \\ \hline
		  	\multicolumn{1}{|c|}{Ara{\'u}jo et al.\cite{araujo2017classification}} & 77.8~ & 72.2~ & 77.8~ & - & - \\ \hline
		\end{tabular}
		\end{center}
	\end{table}
	\vspace{-15px}
To further experiment the effect of using spatial average layer at the end of the patch-wise network, we use the hyper-parameter $C$ that controls the depth of the output feature maps. We examined different values of $C$ and measured the accuracy using our validation set (Table \ref{tab:res}). In our tests, the network with only one output channel outperforms the others corroborating the idea that having many filters for a single concept impose extra burden on the next network, which needs to adjust with all variations from the previous network \cite{lin2013network}. The ROC curves for $C=1$ and $C=4$ are illustrated in Figure \ref{fig:roc}.
\begin{table}[!htb]
		\begin{center}
		\caption{Detailed results of both networks over different classes} 
		\label{tab:res}
		\def\arraystretch{1.12}
		\begin{tabular}{|c|c|c c |c|c|c c c|c|}
			\hline
    		& &\multicolumn{3}{c|}{Patch-wise} &  & \multicolumn{4}{c|}{Image-wise}\\
	  		\hline
	 		Feature maps& Class & Precision & Recall &  Accuracy & Class & Precision & Recall & AUC& Accuracy\\
	 		\hline
			\multirow{4}{*}{$64 \times 64 \times 1$} & Normal & 0.80 & 0.81& \multirow{4}{*}{0.82} & Normal & 0.95 & 1.00 & 0.99 &\multirow{4}{*}{\textbf{0.94}} \\ \cline{2-4}\cline{6-9}
	 		& Benign & 0.81 & 0.77&  & Benign & 0.95 & 0.95& 0.98 &\\ \cline{2-4}\cline{6-9}
	 		& InSitu & 0.76 & 0.85 &  & InSitu & 0.86 & 0.95& 0.99 &\\ \cline{2-4}\cline{6-9}
	 		& Invasive & 0.93& 0.86& & Invasive & 1.00 & 0.85& 0.98&\\ \hline
			\multirow{4}{*}{$64 \times 64 \times 3$} & Normal & 0.81 & 0.86& \multirow{4}{*}{0.86} & Normal & 0.94 & 0.85& 0.97 &\multirow{4}{*}{0.90} \\ \cline{2-4}\cline{6-9}
			& Benign & 0.85 & 0.79&  & Benign & 0.86 & 0.90& 0.96&\\ \cline{2-4}\cline{6-9}
			& InSitu & 0.86 & 0.88 &  & InSitu & 0.83 & 0.95& 0.98&\\ \cline{2-4}\cline{6-9}
			& Invasive & 0.93& 0.90& & Invasive & 1.00 & 0.90& 0.98&\\ \hline
			\multirow{4}{*}{$64 \times 64 \times 4$} & Normal & 0.81 & 0.90& \multirow{4}{*}{0.86} & Normal & 0.79 & 0.95 & 0.98 &\multirow{4}{*}{0.88} \\ \cline{2-4}\cline{6-9}
			& Benign & 0.85 & 0.79&  & Benign & 1.00 & 0.70& 0.97&\\ \cline{2-4}\cline{6-9}
			& InSitu & 0.90 & 0.84 &  & InSitu & 0.86 & 0.90& 0.99&\\ \cline{2-4}\cline{6-9}
			& Invasive & 0.90& 0.93& & Invasive &  0.90& 0.95& 0.97&\\ \hline
			\multirow{4}{*}{$64 \times 64 \times 16$} & Normal & 0.77 & 0.86& \multirow{4}{*}{0.82} & Normal & 0.78 & 0.90& 0.98&\multirow{4}{*}{0.85} \\ \cline{2-4}\cline{6-9}
			& Benign & 0.85 & 0.68&  & Benign & 0.83 & 0.75& 0.96&\\ \cline{2-4}\cline{6-9}
			& InSitu & 0.80 & 0.87 &  & InSitu & 0.94 & 0.80& 0.96&\\ \cline{2-4}\cline{6-9}
			& Invasive & 0.86& 0.85& & Invasive & 0.86 & 0.95& 0.99&\\ \hline
		\end{tabular}
		\end{center}
	\end{table}

%% file: sections/conclusion.tex
	In this manuscript, we considered the problem of breast cancer classification using microscopy tissue images. We utilized deep learning techniques and proposed a novel two-stage CNN pipeline to overcome the hardware limitations imposed by processing of very large images. The first so called \textsl{patch-wise} network acts on the smaller patches of the whole image and outputs spatially smaller feature maps. The second network is performing on top of the patch-wise network. It receives stack of feature maps from the patch-wise network as input and generates image-level label scores. In this framework, patch-wise network is responsible for capturing the local features of the input while the image-wise network is learning to combine those features and find the relationship between neighbouring patches to globally infer characteristics of the image and generate class confident scores. The main contribution of this work is presenting a pipeline which is able to process large scale images using minimal hardware. 
	
	We trained the networks using the ICIAR 2018 grand challenge on BreAst Cancer Histology (BACH) dataset. The proposed method yields $95\%$ accuracy on the four-class validation set compared to previously reported $77\%$ accuracy rates in \cite{araujo2017classification}. It is worth noting that inference time of our scheme is in the order of milliseconds.  The pre-trained weights of our networks are relatively small in size ($7.9$MB patch-wise, $1.6$MB image-wise) compared to other state-of-the-art networks (hundreds of MB) which makes them suitable for practical settings.

	We trained two networks separately on the same labels with the same loss function. One might fairly argue that training the patch-wise network with the same labels as the image-wise network is a disadvantage to the performance of our model. Clearly, not every patch in an image represents the same category. 

	One alternative would be to train both networks end-to-end using only one loss function that back-propagates through both networks. In this scheme, both networks are interconnected to let the flow of gradient and therefore cost is minimized by updating both networks' parameters together. 	In our experiments, we found that such model requires a very large memory footprint that makes it impractical to apply in practice. We plan to further investigate this framework in the future and focus on its improvements.

%% file: bach.bbl
\begin{thebibliography}{10}

\bibitem{siegel2016cancer}
Siegel, R.L., Miller, K.D., Jemal, A.:
\newblock Cancer statistics, 2016.
\newblock CA: a cancer journal for clinicians \textbf{66}(1) (2016)  7--30

\bibitem{ghaznavi2013digital}
Ghaznavi, F., Evans, A., Madabhushi, A., Feldman, M.:
\newblock Digital imaging in pathology: whole-slide imaging and beyond.
\newblock Annual Review of Pathology: Mechanisms of Disease \textbf{8} (2013)
  331--359

\bibitem{araujo2017classification}
Ara{\'u}jo, T., Aresta, G., Castro, E., Rouco, J., Aguiar, P., Eloy, C.,
  Pol{\'o}nia, A., Campilho, A.:
\newblock Classification of breast cancer histology images using convolutional
  neural networks.
\newblock PloS one \textbf{12}(6) (2017)  e0177544

\bibitem{hou2016patch}
Hou, L., Samaras, D., Kurc, T.M., Gao, Y., Davis, J.E., Saltz, J.H.:
\newblock Patch-based convolutional neural network for whole slide tissue image
  classification.
\newblock In: Proceedings of the IEEE Conference on Computer Vision and Pattern
  Recognition. (2016)  2424--2433

\bibitem{cruz2014automatic}
Cruz-Roa, A., Basavanhally, A., Gonz{\'a}lez, F., Gilmore, H., Feldman, M.,
  Ganesan, S., Shih, N., Tomaszewski, J., Madabhushi, A.:
\newblock Automatic detection of invasive ductal carcinoma in whole slide
  images with convolutional neural networks.
\newblock In: SPIE medical imaging. Volume 9041., International Society for
  Optics and Photonics (2014)  904103--904103

\bibitem{iciar2018}
15th International Conference~on Image~Analysis, Recognition:
\newblock {I}{C}{I}{A}{R} 2018 grand challenge
  \url{https://iciar2018-challenge.grand-challenge.org/}.

\bibitem{kowal2013computer}
Kowal, M., Filipczuk, P., Obuchowicz, A., Korbicz, J., Monczak, R.:
\newblock Computer-aided diagnosis of breast cancer based on fine needle biopsy
  microscopic images.
\newblock Computers in biology and medicine \textbf{43}(10) (2013)  1563--1572

\bibitem{george2014remote}
George, Y.M., Zayed, H.H., Roushdy, M.I., Elbagoury, B.M.:
\newblock Remote computer-aided breast cancer detection and diagnosis system
  based on cytological images.
\newblock IEEE Systems Journal \textbf{8}(3) (2014)  949--964

\bibitem{filipczuk2013computer}
Filipczuk, P., Fevens, T., Krzyzak, A., Monczak, R.:
\newblock Computer-aided breast cancer diagnosis based on the analysis of
  cytological images of fine needle biopsies.
\newblock IEEE Transactions on Medical Imaging \textbf{32}(12) (2013)
  2169--2178

\bibitem{brook2006breast}
Brook, A., El-Yaniv, R., Isler, E., Kimmel, R., Meir, R., Peleg, D.:
\newblock Breast cancer diagnosis from biopsy images using generic features and
  svms.
\newblock IEEE Transactions on Information Technology in Biomedicine (2006)

\bibitem{zhang2011breast}
Zhang, B.:
\newblock Breast cancer diagnosis from biopsy images by serial fusion of random
  subspace ensembles.
\newblock In: Biomedical Engineering and Informatics (BMEI), 2011 4th
  International Conference on. Volume~1., IEEE (2011)  180--186

\bibitem{cirecsan2013mitosis}
Cire{\c{s}}an, D.C., Giusti, A., Gambardella, L.M., Schmidhuber, J.:
\newblock Mitosis detection in breast cancer histology images with deep neural
  networks.
\newblock In: International Conference on Medical Image Computing and
  Computer-assisted Intervention, Springer (2013)  411--418

\bibitem{simonyan2014very}
Simonyan, K., Zisserman, A.:
\newblock Very deep convolutional networks for large-scale image recognition.
\newblock arXiv preprint arXiv:1409.1556 (2014)

\bibitem{ioffe2015batch}
Ioffe, S., Szegedy, C.:
\newblock Batch normalization: Accelerating deep network training by reducing
  internal covariate shift.
\newblock In: International Conference on Machine Learning. (2015)  448--456

\bibitem{maas2013rectifier}
Maas, A.L., Hannun, A.Y., Ng, A.Y.:
\newblock Rectifier nonlinearities improve neural network acoustic models.
\newblock In: Proc. ICML. Volume~30. (2013)

\bibitem{springenberg2014striving}
Springenberg, J.T., Dosovitskiy, A., Brox, T., Riedmiller, M.:
\newblock Striving for simplicity: The all convolutional net.
\newblock arXiv preprint arXiv:1412.6806 (2014)

\bibitem{lin2013network}
Lin, M., Chen, Q., Yan, S.:
\newblock Network in network.
\newblock arXiv preprint:1312.4400 (2013)

\bibitem{srivastava2014dropout}
Srivastava, N., Hinton, G.E., Krizhevsky, A., Sutskever, I., Salakhutdinov, R.:
\newblock Dropout: a simple way to prevent neural networks from overfitting.
\newblock Journal of machine learning research \textbf{15}(1) (2014)
  1929--1958

\bibitem{luo2016understanding}
Luo, W., Li, Y., Urtasun, R., Zemel, R.:
\newblock Understanding the effective receptive field in deep convolutional
  neural networks.
\newblock In: Advances in Neural Information Processing Systems. (2016)
  4898--4906

\bibitem{liu2017detecting}
Liu, Y., Gadepalli, K., Norouzi, M., Dahl, G.E., Kohlberger, T., Boyko, A.,
  Venugopalan, S., Timofeev, A., Nelson, P.Q., Corrado, G.S.,  et~al.:
\newblock Detecting cancer metastases on gigapixel pathology images.
\newblock arXiv preprint arXiv:1703.02442 (2017)

\bibitem{macenko2009method}
Macenko, M., Niethammer, M., Marron, J., Borland, D., Woosley, J.T., Guan, X.,
  Schmitt, C., Thomas, N.E.:
\newblock A method for normalizing histology slides for quantitative analysis.
\newblock In: Biomedical Imaging: From Nano to Macro, 2009. ISBI'09. IEEE
  International Symposium on, IEEE (2009)  1107--1110

\bibitem{kingma2014adam}
Kingma, D., Ba, J.:
\newblock Adam: A method for stochastic optimization.
\newblock arXiv preprint arXiv:1412.6980 (2014)

\end{thebibliography}
